\documentclass[conference]{IEEEtran}

\usepackage{algorithm}
\usepackage{algorithmic}
\usepackage{amsmath}
\usepackage{amsfonts}
\usepackage[pdftex]{graphicx}
\usepackage{mathtools}
\usepackage{tikz}
\usepackage{url}
\usepackage{subcaption}
\usepackage{graphicx}
\usepackage{amsthm}
\DeclareMathOperator*{\argmax}{arg\,max}
\DeclareMathOperator*{\argmin}{arg\,min}

\newcommand{\R}{\mathbb{R}}                      
\newcommand{\proj}{\text{Pr}}                    

\newtheorem{thm}{Theorem}
\theoremstyle{definition}

\newcommand\copyrighttext{%
  \footnotesize \copyright 2018 IEEE. Personal use of this material is permitted. Permission from IEEE must be
obtained for all other uses, in any current or future media, including
reprinting/republishing this material for advertising or promotional purposes, creating new
collective works, for resale or redistribution to servers or lists, or reuse of any copyrighted
component of this work in other works.}
\newcommand\copyrightnoticeb{%
\begin{tikzpicture}[remember picture,overlay]
\node[anchor=south,yshift=10pt] at (current page.south) {\fbox{\parbox{\dimexpr\textwidth-\fboxsep-\fboxrule\relax}{\copyrighttext}}};
\end{tikzpicture}%
}

\begin{document}
%
\title{A GPU-Oriented Algorithm Design for Secant-Based Dimensionality Reduction}

\author{\IEEEauthorblockN{Henry Kvinge, Elin Farnell, Michael Kirby, and Chris Peterson}
\IEEEauthorblockA{Department of Mathematics\\
Colorado State University\\Fort Collins, CO 80523-1874}}

\maketitle

\copyrightnoticeb

\begin{abstract}
Dimensionality-reduction techniques are a fundamental tool for extracting useful information from high-dimensional data sets. Because secant sets encode manifold geometry, they are a useful tool for designing meaningful data-reduction algorithms. In one such approach, the goal is to construct a projection that maximally avoids secant directions and hence ensures that distinct data points are not mapped too close together in the reduced space. This type of algorithm is based on a mathematical framework inspired by the constructive proof of Whitney's embedding theorem from differential topology. Computing all (unit) secants for a set of points is by nature computationally expensive, thus opening the door for exploitation of GPU architecture for achieving fast versions of these algorithms. We present a polynomial-time data-reduction algorithm that produces a meaningful low-dimensional representation of a data set by iteratively constructing improved projections within the framework described above. Key to our algorithm design and implementation is the use of GPUs which, among other things, minimizes the computational time required for the calculation of all secant lines. One goal of this report is to share ideas with GPU experts and to discuss a class of mathematical algorithms that may be of interest to the broader GPU community. 
\end{abstract}


\section{Introduction}
High performance computing architectures and massive data centers have created a modern data challenge. Supercomputers generate data at ever more amazing rates, in extreme cases upwards of 20,000,000 cores processing power and processing speeds approaching 100 petaflops.  Also, advances in
data acquisition have us in uncharted territory; e.g., the Australian Square Kilometre Array Pathfinder is generating data measured in petabytes per hour.  

How does one begin to interrogate data sets of this magnitude? In this paper we propose that one of the first questions one can pose about data is mathematical, i.e., what is the dimension of the data? Are fast dimension estimates of massive data sets feasible? What is the role of high-performance computing and parallel architectures for dimension-estimation algorithms?  Examining all points, or large subsets of
points, to identify optimal representation subspaces appears inherently parallelizable. The dimension question provides a window into the complexity of the data and opportunities for data reduction.

Even the question of monitoring these high-performance computing
systems is a challenge.  Mathematical models for anomaly detection provide an interesting direction, but these approaches also suffer from the curse of dimensionality~\cite{Th10,wang2015identity}. Characterizing the behavior of a 100,000,000 variable dynamical system requires new modeling strategies. Again, understanding the dimension of the data may provide the first step to making the modeling procedure tractable~\cite{BK05}.

The outline of this paper is as follows: in Section \ref{sec:overview}, we discuss our proposal for an algorithm that provides both a meaningful projection for dimensionality reduction and a means of estimating the dimension of a data set. We cover foundational mathematical background in Section \ref{sec:background}. We present the specifics of the proposed algorithm in Section \ref{sect-algo} and we demonstrate its use on a synthetic example in Section \ref{sec:trigmoment}. In Section \ref{section-dim-testing}, we use the algorithm to estimate the dimension of three data sets: a synthetic data set, a data set of digital images, and a hyperspectral data set. Finally, in Section \ref{Sect-Noise} we suggest a modified version of the algorithm for noisy data.

\section{Overview}
\label{sec:overview}
High-dimensional data sets can be both a computational burden and difficult to analyze. Data-reduction algorithms offer a way to reduce these difficulties by mapping the data set into a lower-dimensional space with the goal of retaining as much information as possible. A classical example of such an algorithm is principal component analysis (PCA). In \cite{BK05,BK00,BK01}, Broomhead and Kirby developed a new framework for data reduction based on Whitney's embedding theorem, a theorem from differential topology which gives an upper bound on the dimension of Euclidean space required to smoothly embed a compact $k$-dimensional manifold \cite[Section 1.8]{GP10}. The goal of this framework is to produce projections which not only retain differential structure but also have a well-conditioned inverse (roughly, an inverse for which small changes in the domain produce similarly small changes in the range). This property, not necessarily found for projections obtained from other popular methods such as PCA, means that the projection provides a method to compress and decompress the data without loss of information.

In practice, producing projections for a data set $X \subset \R^n$ into $\R^m$ for $m < n$ within the Whitney reduction framework involves finding projections $P: X \subset \R^n \rightarrow \R^m$ for which the smallest value of 
\begin{equation*}
\frac{||P(x_1) - P(x_2)||_{\ell_2}}{||x_1 - x_2||_{\ell_2}}
\end{equation*}
for $x_1, x_2 \in X$ is maximized. This can be more concisely described as follows: let $S$ be the set of all secants of $X$ (that is, the set of differences of all distinct points in $X$). Then we seek a projection $P$ that best preserves $S$. In Section \ref{sect-algo} of this paper we describe a new algorithm to accomplish this, which we call Secant-Avoidance Projection (SAP).

Because the number of secants for a data set of size $n$ is $n(n-1)/2$, even relatively small data sets can have large corresponding secant sets. For this reason, the use of GPU architecture is key to the design of this algorithm. Calculation of the secant set itself can be cumbersome on a CPU as the size of $n$ becomes large. Furthermore, many of the steps in the algorithm involve small independent calculations with each secant in $S$. These can be trivially parallelized and implemented efficiently on a GPU.

We note that our algorithm also provides a method of approximating the dimension of a data set. We outline how this is done in Section \ref{section-dim-testing} and provide applications in this context.

\section{Background Theory}
\label{sec:background}
In this section we discuss some of the 
background related to the practical question of dimension estimation~\cite{hurewicz1948dimension}.

\subsection{Topological Dimension}
For data sampled from smooth manifolds, e.g, intervals, circles, spheres and tori,
the space looks locally flat and it is
only at larger scales that the curvature
becomes apparent.  The dimension
of this locally flat space, i.e., 
the tangent space, is taken to be the dimension of the manifold (which is the same at every point on the manifold).
Following \cite{broomhead_jones_king87}, we introduce
the $\epsilon$-tangent space determined
by the singular vectors associated with the singular values of the singular
value decomposition (SVD) that scale linearly
for small $\epsilon$.  It is this
subset of the basis
elements that determine
the tangent space and the topological
dimension of the manifold~\cite{munkres75}.

\begin{thm} \cite{BINR91}
The basis vectors whose singular spectrum scales as $\epsilon$ for small $\epsilon$ form a basis for the tangent space centered at $p$ of an $m$-dimensional manifold. i.e., those basis vectors for which the set of singular values $\left\{\sigma_i\right\}$ of an $\epsilon$-ball centered at $p$ have
$$\sigma_i(p, \epsilon) = k \epsilon+O(\epsilon^2),\quad 1\le i \le m$$ for some constant $k$ form a basis for the tangent space at $p.$
\end{thm}

It is implicit in the above statement that any basis vector of the local SVD that does not scale linearly lives in a space perpendicular to the tangent space. This is a tractable computation to approximate dimension, especially in low dimensions; see, e.g., \cite{broomhead_97a,kirby_hundley1999}. However,  in theory it requires the computation of SVDs over many scales at each point on the data set.

\subsection{Whitney's Theorem}
A foundational result in differential topology, Whitney's embedding theorem gives an upper bound on the dimension of Euclidean space needed to smoothly embed a finite-dimensional compact differentiable manifold. Recall that an \emph{immersion} is a differentiable function $f:M \rightarrow N$ between two differentiable manifolds $M$ and $N$, whose derivative is everywhere injective.

\begin{thm} \cite{GP10}
Let $M$ be an $m$-dimensional differentiable compact manifold. Then there exists an embedding into $\R^{2m+1}$.
\end{thm}

The proof of this theorem in \cite{Hir94} uses the construction of the secant
set $S$ and argues that the 
projection has an inverse
as long as the point of projection does not lie on or infinitesimally close to any secant line of $M$. 

\subsection{Bi-Lipschitz Criterion}
A function $f(x)$ is said to be {\it bi-Lipschitz} on $S$ if for all $x,y \in S$ it
holds that
\begin{equation} \label{eqn-bilipschitz}
m_1 \| x - y \|_{\ell_2} \le \| f(x) - f(y) \|_{\ell_2} \le m_2 \| x - y \|_{\ell_2}.
\end{equation}
The constants $m_1, m_2$ can be interpreted as follows:
\begin{itemize}
\item $m_1$ is the injectivity/immersivity parameter and prevents pairs of points from collapsing,
\item $m_2$ is the Lipschitz constant and prevents pairs of points from blowing apart.
\end{itemize}

The characterization of a function as bi-Lipschitz is connected to 
dimension estimation~\cite{AnHuKi02,kirby_wiley2}.

\begin{thm} \cite{Fal03}
If a function $f:S \rightarrow T$ is bi-Lipschitz, then
$$\dim(S) =  \dim(T),$$ 
where the dimension can be taken
as the topological dimension, or the Hausdorff dimension.
\end{thm}

\begin{thm}\cite{BK00}
If a function $f(x)$ is bi-Lipschitz, then the inverse function $f^{-1}$
is also bi-Lipschitz with injectivity parameter $1/m_2$ and Lipschitz constant $1/m_1$, i.e., 
\begin{equation*}
\frac{1}{m_2} \| x' - y' \|_{\ell_2} \le \| f^{-1}(x') - f^{-1}(y') \|_{\ell_2} \le \frac{1}{m_1} \| x' - y' \|_{\ell_2}.
\end{equation*}
\end{thm}
Typically reduction mappings are not optimized for reconstruction and may 
have ill-conditioned inverses. We note that when $f$ is a projection, the upper bound in \eqref{eqn-bilipschitz} is automatically satisfied and only the lower bound needs to be checked.

\section{The Secant-Avoidance Projection  algorithm} \label{sect-algo}

In this section we describe our algorithm for data reduction. Given a data set of $k$ points $X = \{x^{(1)}, \dots, x^{(k)}\} \subset \R^{n}$ we want to construct a projection $\proj: \R^n \rightarrow \R^m$ such that the distances between points in $X$ are preserved. To accomplish this, we propose, for all pairs $x^{(i)}, x^{(j)} \in X,$ to calculate and store the corresponding normalized secant 
\begin{equation*}
s := \frac{{x^{(i)}- x^{(j)}}}{||x^{(i)}- x^{(j)}||_{\ell_2}}.
\end{equation*}
Note that we normalize each of these secants so that when we compare their projections, those corresponding to close points and far points are on an equal footing. The calculation of all secants for a data set is trivially parallelizable and is an ideal task for a GPU. 

We store secants as the columns of an $n \times p$ matrix $S$ and for iteration $i$ of our algorithm we realize our current projection $\proj^{(i)}$ via an $n \times m$ matrix $P^{(i)}$ whose columns are  orthonormal vectors spanning an $m$-dimensional subspace $M \subset \R^n$. We write these column vectors as $p_1^{(i)}, p_2^{(i)}, \dots, p_m^{(i)}$, and $P^{(i)} = [\;p_1^{(i)} \;|\; p_2^{(i)} \;|\cdots|\; p_m^{(i)}\;]$. The projection $\proj^{(i)}(s)$ of secant $s$ can then be computed as ${P^{(i)}}^T s$. Our algorithm is designed to solve the optimization problem
\begin{equation*}
P^* = \argmax_P\Big( \min_{s \in S} ||P^T s||_{\ell_2}\Big)
\end{equation*} 
where the maximum is taken over all projections $P$ from $\R^n$ to an $m$-dimensional subspace.

As our initial projection we choose $P^{(1)}=[\;p_1^{(1)}\;|\; p_2^{(1)}\;|\; \cdots \;|\; p_m^{(1)}\;]$ to be the first $m$ columns of $U$, where $S = U\Sigma V^T$ is the singular value decomposition of $S.$ In other words, $\proj^{(1)}$ is the projection given by PCA. 

After initializing $P^{(1)}$, the algorithm proceeds as follows: at each iteration $i$ we calculate the $\ell_2$-norms of all secants under the current projection. In other words we find the column of ${P^{(i)}}^T  S$ with the smallest $\ell_2$ norm; call this column index $j^*.$ The column index $j^*$ corresponds to the index of the secant $s_{j^*}$ as a column in $S$ that is least well preserved by $\proj^{(i)}$. 

We construct $P^{(i+1)}$ from $P^{(i)}$ by rotating $P^{(i)}$ by a small amount toward the direction specified by the secant $s_{j^*}.$ In order to shift $P^{(i)}$ toward this secant we first calculate the projection of this secant $P^{(i)}{P^{(i)}}^T  s_{j^*}$ in $\R^n$. Let $p_t^{(i)}$ be the vector from $p_1^{(i)}, \dots, p_m^{(i)}$ which maximizes $|\langle p_k^{(i)}, s_{j^*} \rangle|$. If all these quantities are zero then we pick $p_t^{(i)}=p_1^{(i)}$. After selecting $p_t^{(i)},$ we run the modified Gram-Schmidt algorithm on the $m$ ordered vectors $\left\{P^{(i)}{P^{(i)}}^T s_{j^*}, p_1^{(i)}, p_2^{(i)}, \dots, p_{t-1}^{(i)}, p_{t+1}^{(i)},\dots, p_m^{(i)}\right\}$ to obtain a new set of orthonormal vectors $$\left\{\frac{P^{(i)}{P^{(i)}}^T s_{j^*}}{\|P^{(i)}{P^{(i)}}^T s_{j^*}\|_{\ell_2}}, p_2^{(i+1)}, p_3^{(i+1)}, \dots, p_{m}^{(i+1)}\right\}$$ which has the same span but now necessarily contains the normalized projection of $s_{j^*}.$ We then set $p_1^{(i+1)}$ to be the unit vector in the direction of $$(1-\alpha) P^{(i)}{P^{(i)}}^T(s_{j^*}) + \alpha(s_{j^*} - P^{(i)}{P^{(i)}}^T(s_{j^*})),$$ where $\alpha \in [0,1]$ is a small constant that controls the amount that our projection shifts at each step. The results presented in this paper were computed using $\alpha=0.01,$ where this value is chosen to ensure that a minor shift is made in each step and to encourage convergence. Experimentally, a value of $\alpha=0.01$ appears to achieve those goals. 

We summarize the Secant-Avoidance Projection algorithm in Algorithm \ref{algo-}. The computational complexity of the SAP algorithm is dominated by the complexity of the SVD and Gram-Schmidt algorithms as well as that of the determination of the secant with the shortest projected norm. A standard implementation approach would result in completion in time $O(n^5).$ However, there are several ways in which one might improve the asymptotic bound. For example, Monte Carlo approaches to the computation of the SVD and smart updates to the lengths of the projected secants stand to improve the time complexity. Such explorations are a topic for future research. 

\begin{algorithm} 
\caption{\label{algo-} Secant-Avoidance Projection}
\begin{algorithmic}[1]
\STATE \textbf{inputs} Secant set $S,$ desired dimension of embedding $m,$ max number of steps (Iterations) or alternative stopping criterion, and shift parameter $\alpha$. 
\STATE Initialize $P^{(1)}$ to be the matrix formed by the first $m$ columns of $U$, where $S = U\Sigma V^T$.
\FOR{$i \leq $ Iterations} 
\STATE Set $s_{j^*} = \argmin_{s \in S}||(P^{(i)})^T s||_{\ell_2}$.
\STATE Set $p_t^{(i)} = \argmax_{1 \leq k \leq m}|\langle p_k^{(i)},s_{j^*}\rangle|$.
\STATE Apply the modified Gram-Schmidt algorithm to $P^{(i)}{P^{(i)}}^Ts_{j^*}, p_1, \dots, p_{t-1}, p_{t+1},\dots, p_m$ to obtain a new orthonormal basis $$\frac{P^{(i)}{P^{(i)}}^Ts_{j^*}}{\|P^{(i)}{P^{(i)}}^Ts_{j^*}\|_{\ell_2}}, p^{(i+1)}_2, \dots, p^{(i+1)}_m.$$
\STATE Set $p_1^{(i+1)}$ to be the unit vector in the direction of $$(1-\alpha) P^{(i)}{P^{(i)}}^T(s_{j^*}) + \alpha(s_{j^*} - P^{(i)}{P^{(i)}}^T(s_{j^*})).$$
\STATE $i+1 \leftarrow i$
\ENDFOR
\RETURN $\proj^{(i)}$
\end{algorithmic}
\end{algorithm}

\section{A synthetic example: trigonometric moment curves}
\label{sec:trigmoment}

In this section we show an example of an application of the SAP algorithm and we compare the results to other common projection methods. We construct a synthetic data set by sampling points from the trigonometric moment curve $\phi: \R \rightarrow \R^{10}$ defined by
\begin{equation*}
\phi(t) := (\cos(t),\sin(t),\cos(2t),\dots, \cos(5t),\sin(5t)).
\end{equation*}

For this example and all following examples, we use a program written in CUDA 8.0 \cite{NBGS08}. On two Nvidia Tesla K80 graphics processing units, we
\begin{itemize}
\item construct the secant set for the relevant data set, 
\item calculate the singular value decomposition for the matrix of all normalized secant vectors using the \textsf{cuSolver} library \cite{cuSolver}.
\item run the algorithm for 100 iterations. For some of the following examples we fix a dimension, and for others, we repeat for various projection dimensions.
\end{itemize}

\begin{figure}
\includegraphics[width=8cm, height=6cm]{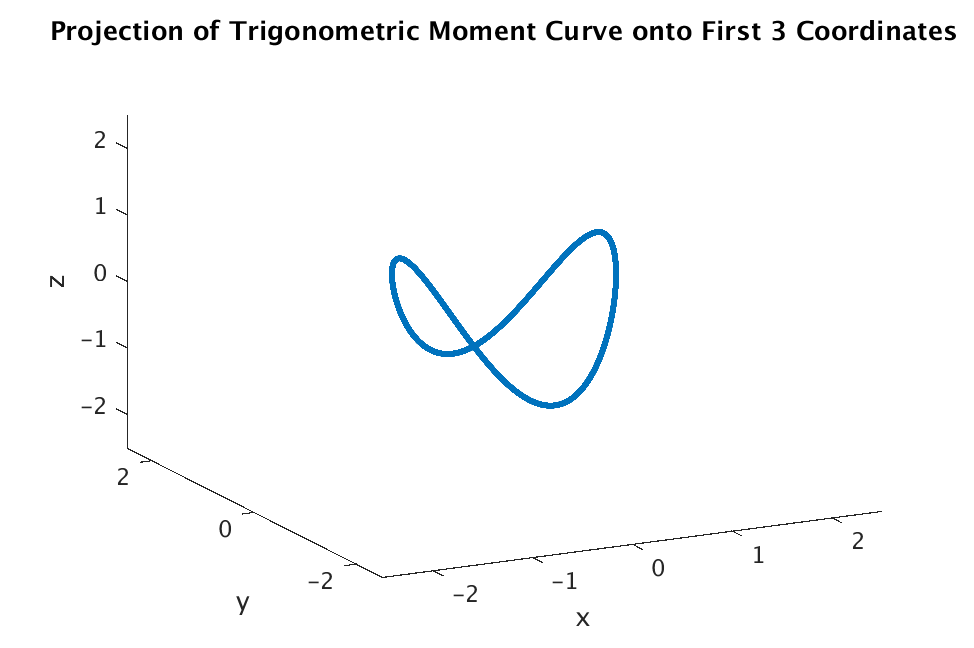}
\caption{\label{figure-proj-onto-3-naive} The projection of the function $\phi$ onto the first three coordinate directions in $\R^{10}$. Clearly, much information is lost by ignoring the content in the remaining 7 coordinates. Compare to the projections into $\mathbb{R}^3$ provided by PCA, which maximizes variance, and to SAP, which seeks a projection with a smooth inverse (Figures \ref{figure-proj-onto-3-PCA} and \ref{figure-proj-onto-3-SAP}, respectively).}
\end{figure}

\begin{figure}
\includegraphics[width=8cm, height=6cm]{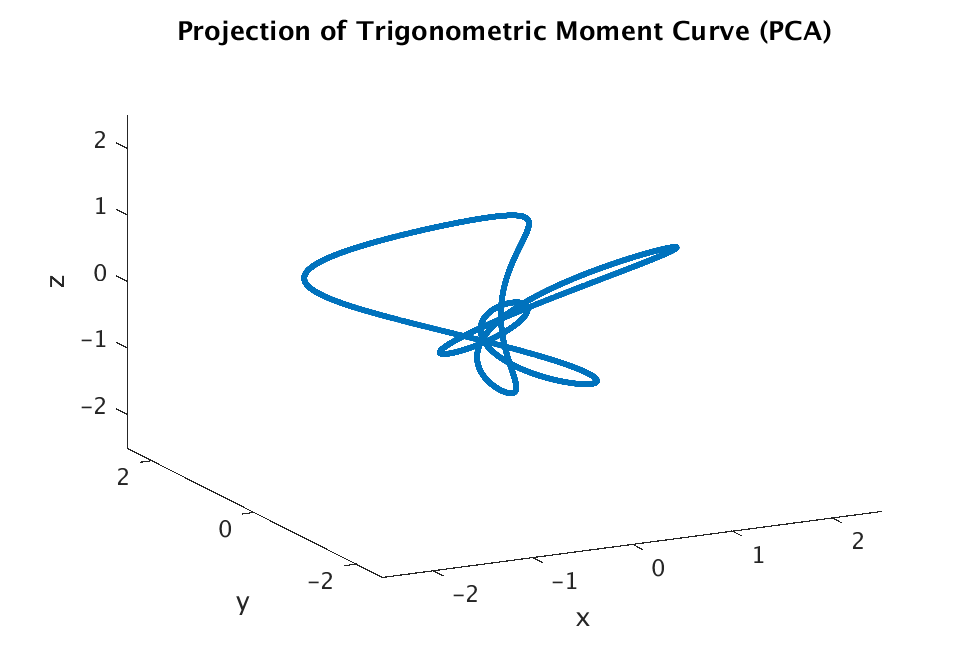}
\caption{\label{figure-proj-onto-3-PCA} The projection of the function $\phi$ into $\mathbb{R}^3$ via PCA. Recall that the projection provided by PCA is the projection that captures maximal variance in the data.}
\end{figure}

\begin{figure}
\includegraphics[width=8cm, height=6cm]{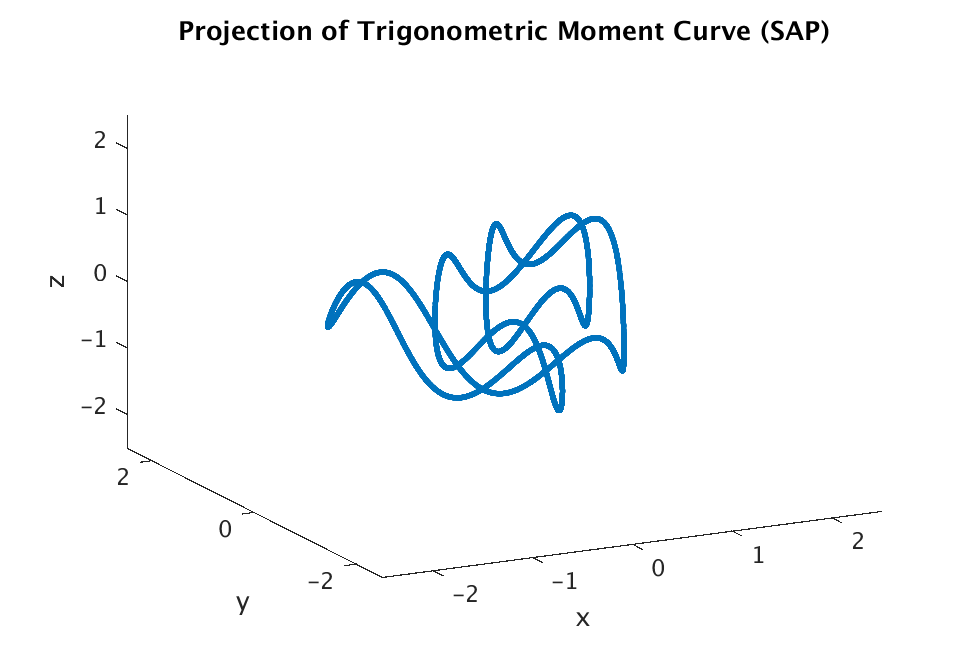}
\caption{\label{figure-proj-onto-3-SAP} The projection of the function $\phi$ into $\mathbb{R}^3$ via SAP. Recall that SAP seeks to produce a projection that maximizes the minimum projected norm across all secants in the data set.}
\end{figure}

Figure \ref{figure-proj-onto-3-naive} shows a projection of the sampled trigonometric moment curve data onto the first three coordinates in $\R^{10}$. Figure \ref{figure-proj-onto-3-PCA} shows the projection into $\R^3$ obtained from PCA (i.e. the projection of the data from $\mathbb{R}^{10}$ to $\mathbb{R}^3$ via $\proj^{(1)}$), and Figure \ref{figure-proj-onto-3-SAP} shows the projection onto $\R^3$ obtained from 100 iterations of the Secant-Avoidance Projection algorithm. In each case, we drew 12,800 values from a uniform random distribution on $[0,2\pi]$ to get a sampling of points on the trigonometric moment curve.

Qualitatively, as seen in Figures \ref{figure-proj-onto-3-naive}, \ref{figure-proj-onto-3-PCA}, and \ref{figure-proj-onto-3-SAP}, the three methods of constructing projections have captured different aspects of the data set through their projections into $\mathbb{R}^3.$ From the perspective of the norm of the shortest projected secant, we find that SAP outperforms both the na\"{i}ve projection and the PCA projection: the norms of the shortest projected secants, in order, are $0.1013,$ $0.0466,$ and $ 0.1677.$  It is worth noting that PCA and SAP are methods of projection defined in terms of particular optimization problems. So while PCA, by construction, will maximize variance, SAP aims to maximize the norm of the shortest projected secant. We see in this example that SAP has indeed outperformed PCA with respect to this metric, as it is designed to do.

\section{Approximating the dimension of a data set} \label{section-dim-testing}

\subsection{A synthetic example} \label{subsection-synthetic-example}

Even in the case where one is not interested in finding a specific projection, the SAP algorithm can be used to better understand the nature of the data. Specifically, it can give a good sense of the dimensionality of the data. One way to observe this is to run the algorithm for a range of different projection dimensions. One should see the norm of the shortest secant for each projection begin to dramatically increase as the projection dimension increases. Via Whitney's theorem, the dimension at which this occurs leads to a good approximation of the dimension of the data.

As a synthetic example we sampled 256 points uniformly from each of the following smooth manifolds:
\begin{enumerate}
\item the curve $f: \R \rightarrow \R^3$
\begin{equation*}
f(t) = (\cos(t), \sin(t), \cos(2t))
\end{equation*}
smoothly embedded into $\R^{15}$,
\item a 2-dimensional torus from $\R^3$ smoothly embedded into $\R^{15}$,
\item a 3-dimensional sphere from $\R^4$ smoothly embedded into $\R^{15}$.
\end{enumerate}

\begin{figure}
\includegraphics[width=8cm]{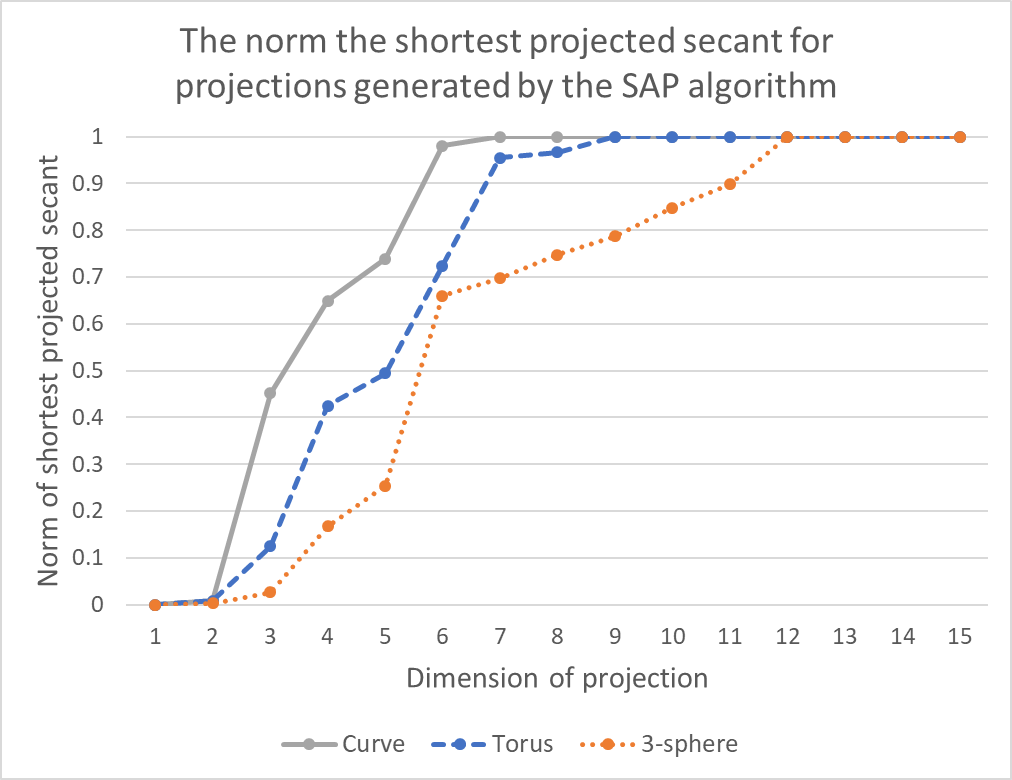}
\caption{\label{figure-plot-dimensionality} The $\ell_2$-norm of the shortest secant projection as a function of the dimension of the projection. Note that the dimensions at which the data sets appear to have successful embeddings agree with those provided as upper bounds by Whitney's embedding theorem (3, 5, and 7, respectively). Also note that the results reflect the relative dimensionality of the data sets.}
\end{figure}

According to Whitney's embedding theorem, we expect that the maximum dimension required to embed the curve, torus, and sphere is 3, 5, and 7, respectively. In Figure \ref{figure-plot-dimensionality}, we see that the SAP algorithm provides an embedding for the three data sets in Euclidean space at these respective dimensions. Specifically, the norm of the shortest projected secant has increased away from zero for these dimensions. We note further that the SAP algorithm provides embeddings at dimensions below the Whitney upper bounds in the case of the torus and $3$-sphere. While it is not uncommon for embeddings to exist in dimensions below the Whitney upper bound, the appearance of such embeddings here further highlights the potential usefulness of the SAP algorithm for dimensionality reduction. We observe as well that there is a qualitative distinction between the behavior of the three data sets shown in Figure \ref{figure-plot-dimensionality}. The norm of the shortest projected secant vector for the three data sets preserves the relative dimensionality relationships between the data sets. 

\subsection{The pumpkin illumination space}
We demonstrate the use of our algorithm on an illumination space data set. We collected the data for use in the Pattern Analysis Lab at Colorado State University. The data consists of images of a solid object under varying illumination conditions. Specifically, we capture images of a plastic Halloween jack-o-lantern under varying illumination conditions. The images are color images of size $480\times 720.$ We pre-process by using PCA to reduce dimension and remove noise. Consequently, we have a set of 200 data points in $\mathbb{R}^{200},$ each of which  corresponds to one of the original images. 

This data set illustrates the potential value of the Secant-Avoidance Projection algorithm. The data set represents a sample of a subspace of the illumination space of the pumpkin, analogous to many existing data sets that capture individuals and objects under varying illumination. Illumination spaces of people have been well-studied and there have been attempts to understand the dimension of such spaces; see, e.g., \cite{BK98,GABK01,CKKPDB07,CBDKKP06}. We conjecture that the illumination space of the pumpkin will have similar characteristics, and the SAP algorithm provides a means of understanding the dimensionality of the sampled subspace of the illumination space (we expect to see a subspace because the variations in illumination were restricted to a subset of those that would provide a broad representation of the set of all possible illuminations).

\begin{figure}
\includegraphics[width=8cm, height=6cm]{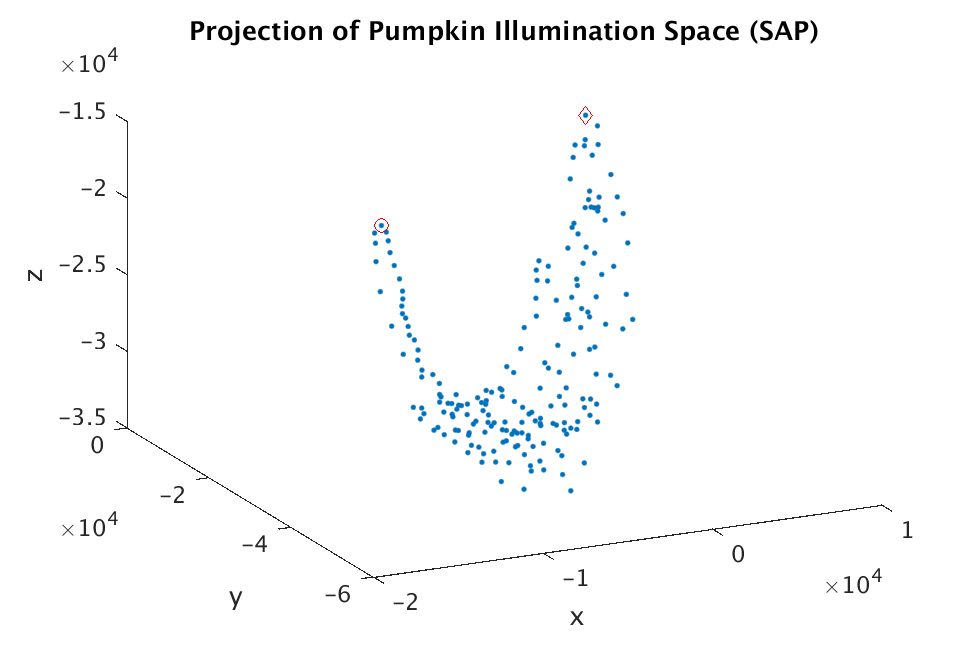}
\caption{\label{figure-projection-pumpkins} The projection of the pumpkin data set using the SAP algorithm. Two distinguished points are marked with a red circle and a red diamond - see Figure \ref{fig:red circle and diamond} for the corresponding images from the original pumpkin illumination space data set.}
\end{figure}
\begin{figure}
\includegraphics[width=8cm, height=6cm]{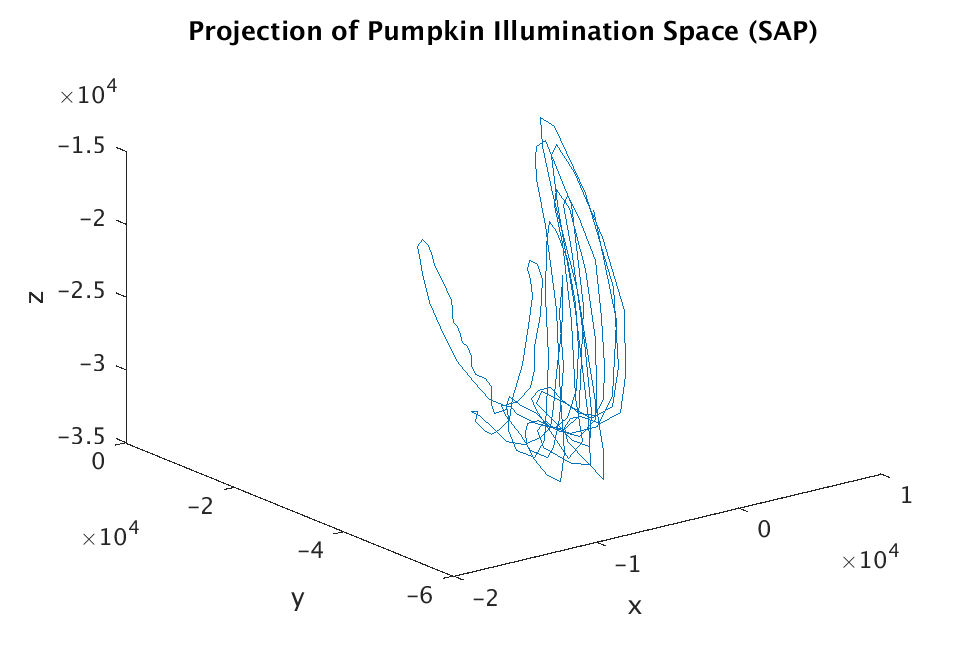}
\caption{\label{figure-projection-pumpkins-connected} The projection of the pumpkin data set using the SAP algorithm with points connected by their order in the time series in which the data was collected.}
\end{figure}

The pumpkin illumination space data is a real-world data set with noise and unknown structure, and we do not know {\em a priori} the natural dimension of the data. But dimensionality, in general, is of primary importance for making data storage and analysis a manageable task. By applying the SAP algorithm, we obtain an estimate of the dimension of this data set. In this case, we note that the SAP algorithm provides a diffeomorphic copy of the data in $\mathbb{R}^3.$ Using Whitney's Theorem, we speculate that the subspace of the illumination space determined by these constrained lighting conditions has dimension one.

\begin{figure}
\centering
\includegraphics[width=9cm, height=4cm]{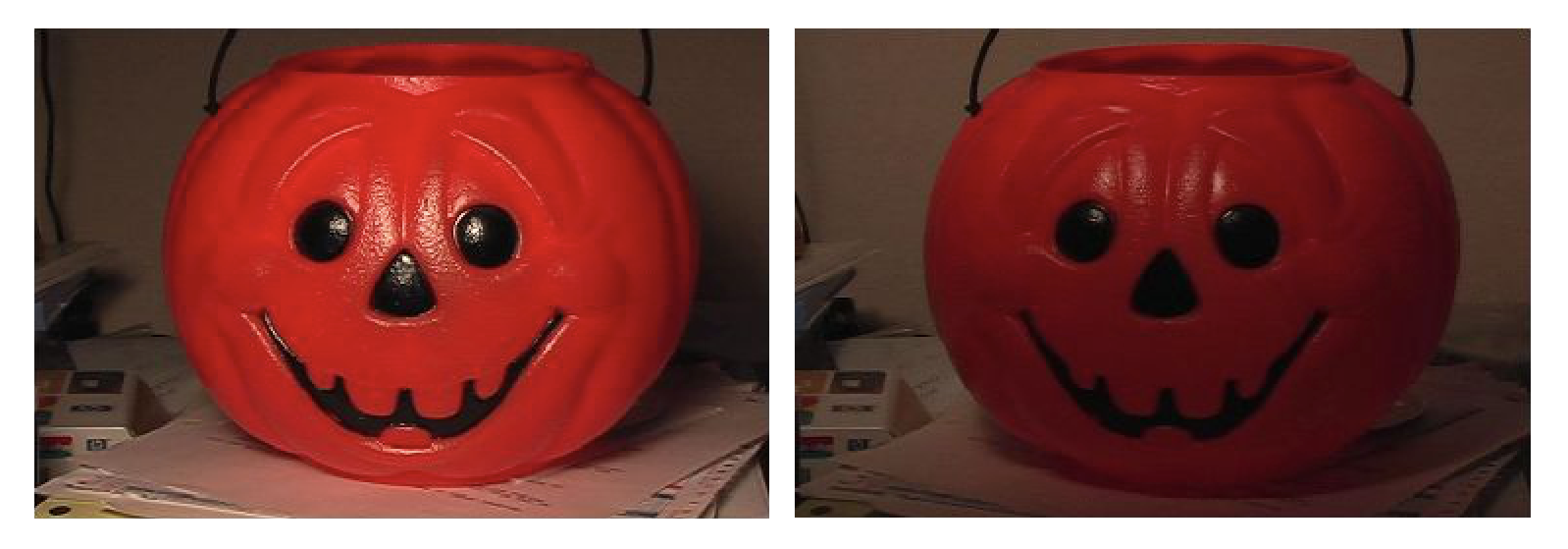}
 \caption{\label{fig:red circle and diamond}Images Corresponding to Extremal Points in Projection to $\mathbb{R}^3$ via SAP. The left image corresponds to the red circle and the right image corresponds to the red diamond in Figure \ref{figure-projection-pumpkins}.} 
\end{figure}
 
We can further understand characteristics of the data set that become apparent after applying the SAP projection. For example, note that two points in the projection shown in Figure \ref{figure-projection-pumpkins} appear to be extrema; by viewing the images from the original data set that correspond to those extremal points, we note that indeed, we have found points that correspond to extreme lighting conditions (Figure \ref{fig:red circle and diamond}). Qualitatively, the SAP projection appears to have preserved fundamental features in the data set. For example, the data set can be viewed as a time series as a consequence of the way the data was captured - each successive data point corresponds to a small change in lighting conditions. Thus, we would hope that a good projection would preserve this feature of the data, and indeed, if one connects the points in the time series as in Figure \ref{figure-projection-pumpkins-connected}, we see that the smooth light variations that varied repeatedly from one extreme to another resulted in an embedding that parametrizes a non-self-intersecting path in three-space. 

\begin{figure}
\centering
\includegraphics[width=8cm, height=6cm]{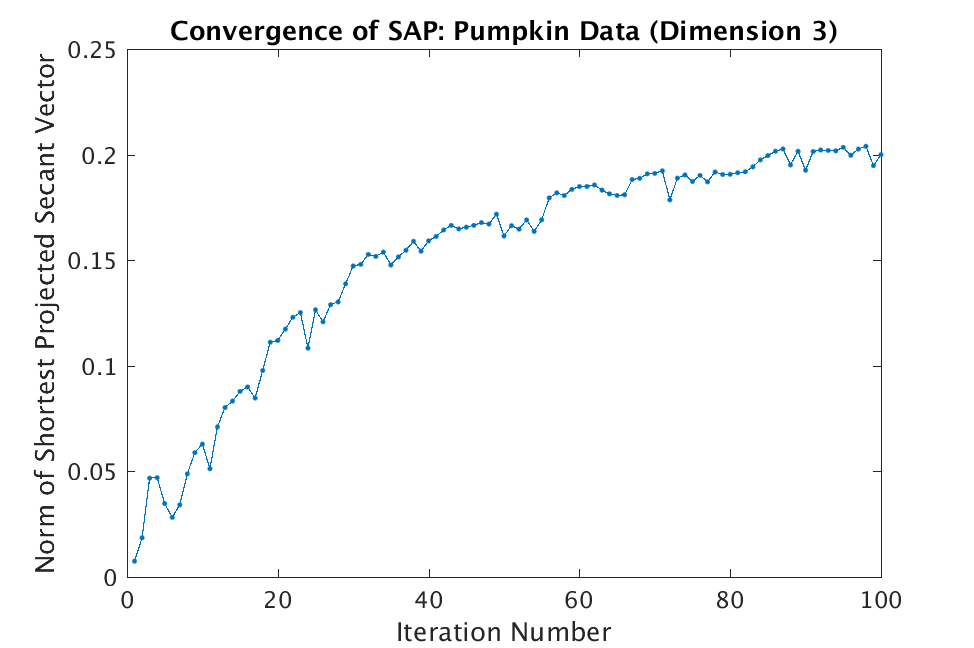}
\caption{\label{figure-projection-pumpkins-convergence} The convergence of the SAP algorithm on the pumpkin data set. The norm of the shortest projected secant vector generally increases with iteration number.}
\end{figure}

In Figure \ref{figure-projection-pumpkins-convergence}, we see the convergence of the algorithm over 100 iterations, where the dimension has been fixed to be three. Note that while there are small perturbations, the norm of the shortest projected secant vector increases with the iteration number and thus we expect that the projections are getting correspondingly better. 

\begin{figure}
\centering
\includegraphics[width=8cm, height=6cm]{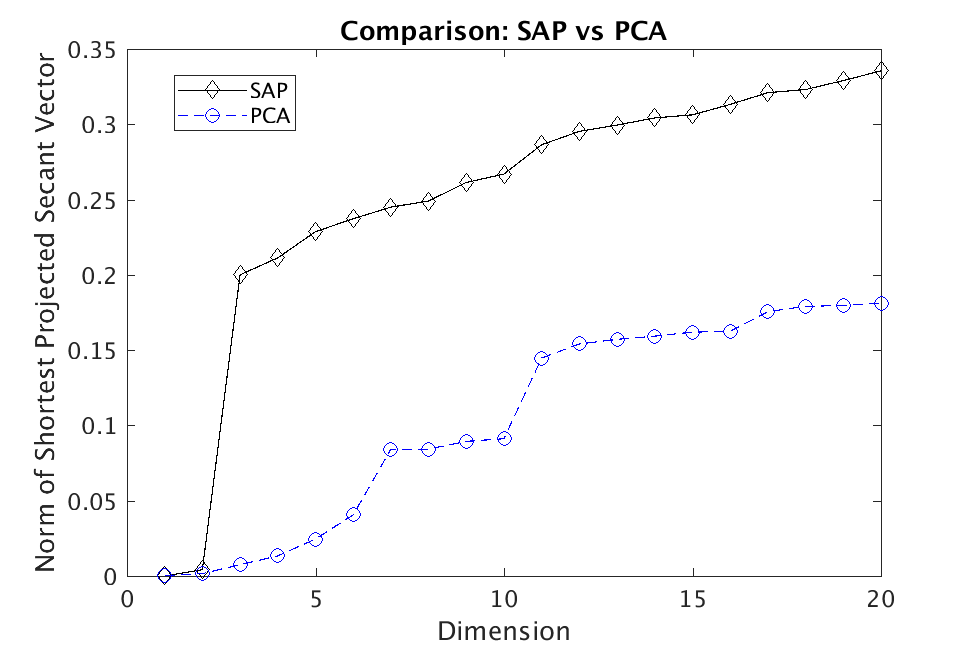}
\caption{\label{figure-projection-pumpkins-comparison} Comparison of the norm of the shortest projected secant vector as a function of dimension after computing a projection using PCA and a projection using SAP on the pumpkin data set. Note that the SAP algorithm outperforms PCA at lower dimensions and persists through high dimensions according to this measure of success.}
\end{figure}

We compare the performance of the SAP algorithm against that of PCA in Figure \ref{figure-projection-pumpkins-comparison}. Note that both would provide relatively good projections into $\mathbb{R}^{20},$ but the SAP algorithm outperforms PCA significantly at lower dimensions and persists in providing a better projection for higher dimensions as well (with respect to this measure of success). Further, the manner of construction of the data set suggests that the data should live on a one-dimensional manifold and hence, the embedding into $\mathbb{R}^3$ by the SAP algorithm is consistent with the upper bound from Whitney's Theorem.

\subsection{Indian Pines hyperspectral data cube}

As our final example we consider the application of our algorithm to a modified version of the Indian Pines hyperspectral data set (some bands covering the region of water absorption are removed) \cite{IP}. The data cube is $145 \times 145 \times 200;$ that is, there are $200$ bands, each with spatial resolution of $145 \times 145$. We define a data set $X\subset\mathbb{R}^{200}$ to be the collection of vectors of spectral information taken across all pixel locations. We then have $|X|=21,025.$ The number of secants associated to this number of points is too massive to handle with the current version of the SAP algorithm so we use a sampling technique and show that in this example at least, such a strategy yields similar results across different random samples.

\begin{figure}
\includegraphics[width=8cm]{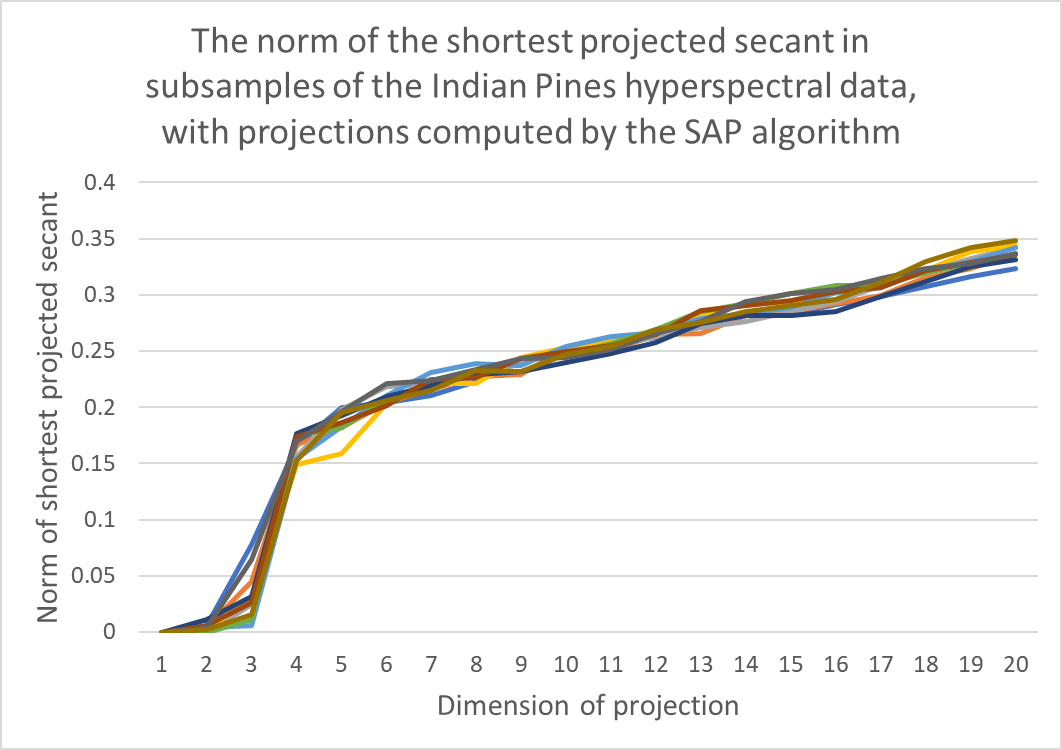}
{\caption{\label{figure-IP-samples} The $\ell_2$-norm of the shortest secant projection as a function of the dimension of the projection for the Indian Pines data set. Each curve in the graph corresponds to SAP-algorithm projections computed for a subsample of 512 points from the Indian Pines data set. Note the sudden jump in $\ell_2$-norm when the dimension of the projection increases to $4$.}}
\end{figure}

In Figure \ref{figure-IP-samples} we show the plot of the dimension of the projection versus the norm of the shortest projected secant for ten experiments in which we randomly sample $512$ points from $X$ and run the SAP algorithm on the associated secant set. As can be seen, aside from minor deviations, these curves are all very similar suggesting that sampling points might be a reasonable strategy to adopt in the setting when the total number of secants in the data set is very large. Note also that in all cases the curve jumps from less than $0.05$ to $0.2$ as the projection dimension increases from $3$ to $4$. Thus, we have an embedding of the Indian Pines data into Euclidean space of dimension 4. If the data resides on a manifold, that manifold most likely has dimension two. 

\section{Data with noise} \label{Sect-Noise}

In the real world, data is almost always contaminated with some amount of noise. If a data set in $\mathbb{R}^n$ sits on an $m$-dimensional manifold $M$ with $m < n$, then adding noise will tend to push data points off of $M,$ thereby increasing its apparent dimension. Since we are usually more interested in calculating the dimension of the pure signal without noise, it is useful to develop tests of dimensionality which are resistant to added noise. In this section we suggest an adaptation to the SAP algorithm for use on noisy data. 

The idea behind this adaptation stems from the simple observation that when noise is added to a data set it perturbs the positions of points. We therefore expect that the direction of secants between points which are close together will change much more than the direction of secants between points which are far apart. In fact, we can imagine that if two points are sufficiently close, then when noise is added, their secant could be rotated to any direction.

We propose to adapt the SAP algorithm for the setting of noisy data by including an additional step in which secants are thresholded by length. That is, prior to normalization, we discard any secants with length less than some predetermined threshold $\ell$. This thresholding value $\ell$ is then an additional parameter to the algorithm. There is a clear trade-off in potential choices of the size of $\ell$: a small $\ell$ may lead to noise-based structure persisting in the data, and a large $\ell$ may cause small-scale structure in the data to be lost.

\begin{figure}
\includegraphics[width=8cm]{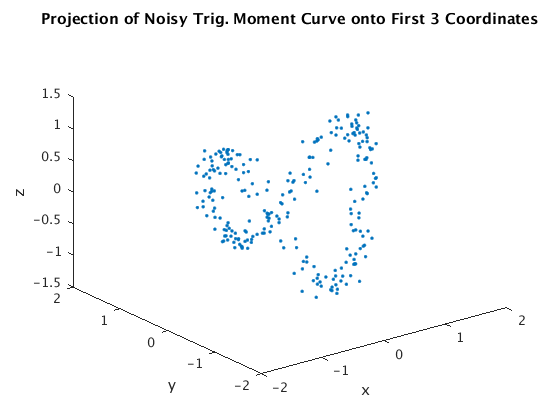}{\caption{\label{NoisyTrig} Projection of Trigonometric Moment Curve with noise onto first three coordinate directions. The noise is independent for each coordinate and has mean 0 and standard deviation 0.1.}}
\end{figure}

\begin{figure}
\includegraphics[width=8cm]{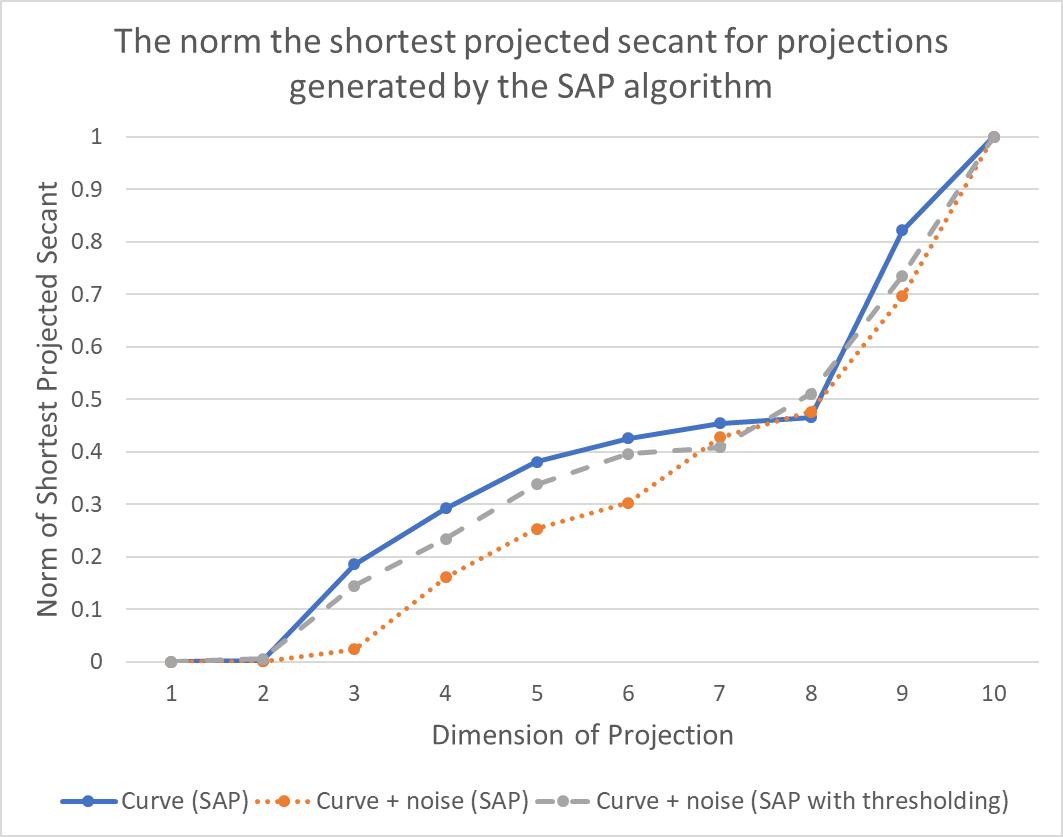}{\caption{\label{Norms-threshold} We compare the norm of the shortest projected secant for: 1) SAP applied to points sampled from the curve defined by $\phi$ from Section \ref{sec:trigmoment}, 2) SAP applied to $\phi$ with noise added, and 3) SAP with thresholding applied to $\phi$ with noise added. Note that the thresholding procedure has improved the embedding and made the results more similar to the no-noise setting.}}
\end{figure}

In Figure \ref{NoisyTrig} we show the result of adding random Gaussian noise to points drawn from the trigonometric moment curve $\phi: \mathbb{R} \rightarrow \mathbb{R}^{10}$ from Section \ref{sec:trigmoment}; the noise is added independently in each coordinate and has mean $0$ and standard deviation $0.1$. Figure \ref{Norms-threshold} shows the result of running the SAP algorithm on the data without any noise added, with the noise added but without thresholding, and with the noise added and thresholding secants at length 2. Note that the thresholding procedure has improved the norm of the shortest projected secant of the noisy data to be more similar to the setting in which there is no noise. This is notably the case for projections of dimensions $3$-$6$, which is where we would look to better understand the dimension of this data set. In fact, comparing Figure \ref{Norms-threshold} to Figure \ref{figure-plot-dimensionality}, we note that  the plot of $\phi$ with noise and no thresholding looks similar to the plot of a genuinely higher dimensional data set. Thresholding secants removes this ambiguity.

\section{Conclusion}
In this paper we described a novel algorithm for dimensionality reduction. The Secant-Avoidance Projection algorithm produces projections of the data into a lower-dimensional space. Importantly, these projections preserve dimension and have smooth inverse. Since our algorithm requires many independent calculations with the elements of the secant set, it is well-suited to a GPU implementation. We also show that besides finding good projections of a data set, our algorithm can be used to identify the approximate dimension of a data set. 

There remain several related open questions and directions for future work; we highlight a selection here.
\begin{enumerate}
\item We have proposed a variation of the SAP algorithm for noisy data. This variation incorporates a threshold $\ell,$ where we discard any secants shorter than $\ell$. A next step is the development of a rigorous method of determining $\ell$ for an arbitrary data set.
\item GPUs provide an ideal setting for the SAP algorithm because of the natural parallelization of computation of the secant set. Further work should be done to develop an efficient algorithm that utilizes GPUs to their full extent throughout the implementation.
\item We recognize that the advantages gained by parallel computation of the secant set on the GPU have limitations. One avenue for further research lies in potential means of combining information from repeated sampling from a data set with independent applications of the SAP algorithm. One algorithm we suggest is to cluster resulting projections and to compute an appropriate average for each cluster, resulting in several potentially useful methods of projection. Such an algorithm would be particularly useful in the setting in which the memory required for computation of the full secant set is infeasible.
\item The proposed algorithm provides no guarantees of convergence to a global optimum. Extensions of this work could propose sufficient conditions for global extrema or an algorithm that does have such a guarantee.
\end{enumerate}

With high-dimensional data being generated at an unprecedented rate, we can expect that demand will only increase for methods of (i) deducing the actual dimension of data and (ii) reducing the dimension of data. Ultimately, we hope that this paper will serve as a first step toward a broader conversation on how to best harness the power of GPU computing to develop secant-based methods that offer efficient solutions to these problems.

\section*{Acknowledgements}
This paper is based on research partially supported by the National Science Foundation
	under Grants No. DMS-1513633, and DMS-1322508 as well as  DARPA awards  N66001-17-2-4020 and D17AP00004.
    
\bibliographystyle{IEEEbib.bst}
\bibliography{references,complete5,KirbyFebruary2018,books}

\begin{thebibliography}{10}

\bibitem{Th10}
Joshua Thompson, David~W. Dreisigmeyer, Terry Jones, Michael Kirby, and Joshua
  Ladd,
\newblock ``Accurate fault prediction of {B}lue{G}ene/{P} {RAS} logs via
  geometric reduction,''
\newblock in {\em Proceedings 1st Workshop on Fault-Tolerance for HPC at
  Extreme Scale (FTXS 2010)}, Chicago, Illinois, 2010.

\bibitem{wang2015identity}
Kun Wang, Josh Thompson, Chris Peterson, and Michael Kirby,
\newblock ``Identity maps and their extensions on parameter spaces:
  Applications to anomaly detection in video,''
\newblock in {\em Science and Information Conference (SAI), 2015}. IEEE, 2015,
  pp. 345--351.

\bibitem{BK05}
D.S. Broomhead and M.~Kirby,
\newblock ``Large dimensionality reduction using secant-based projection
  methods: The induced dynamics in projected systems,''
\newblock {\em Nonlinear Dynamics (Special Issue on Reduced Order Modelling)},
  vol. 41, no. 1-3, pp. 47--67, 2005.

\bibitem{BK00}
D.S. Broomhead and M.~Kirby,
\newblock ``A new approach for dimensionality reduction: Theory and
  algorithms,''
\newblock {\em SIAM J. of Applied Mathematics}, vol. 60, no. 6, pp. 2114--2142,
  2000.

\bibitem{BK01}
D.S. Broomhead and M.~Kirby,
\newblock ``The {W}hitney reduction network: a method for computing
  autoassociative graphs,''
\newblock {\em Neural Computation}, vol. 13, pp. 2595--2616, 2001.

\bibitem{GP10}
Victor Guillemin and Alan Pollack,
\newblock {\em Differential topology},
\newblock AMS Chelsea Publishing, Providence, RI, 2010,
\newblock Reprint of the 1974 original.

\bibitem{hurewicz1948dimension}
W.~Hurewicz and H.~Wallman,
\newblock {\em Dimension theory},
\newblock Princeton mathematical series. Princeton University Press, 1948.

\bibitem{broomhead_jones_king87}
D.~S. Broomhead, R.~Jones, and G.~P. King,
\newblock ``Topological dimension and local coordinates from time series
  data,''
\newblock {\em J. Phys. A: Math. Gen}, vol. 20, pp. L563--L569, 1987.

\bibitem{munkres75}
James~R. Munkres,
\newblock {\em Topology: a first course},
\newblock Prentice Hall, Englewood Cliffs, N.J., 1975.

\bibitem{BINR91}
DS~Broomhead, R~Indik, AC~Newell, and DA~Rand,
\newblock ``Local adaptive {G}alerkin bases for large-dimensional dynamical
  systems,''
\newblock {\em Nonlinearity}, vol. 4, no. 2, pp. 159, 1991.

\bibitem{broomhead_97a}
R.V. Abadi, D.~S. Broomhead, R.A. Clement, J.P. Whittle, and R.~Worfolk,
\newblock ``Dynamical systems analysis: A new method of analysing congenital
  nystagmus waveforms,''
\newblock Technical Report 97-3, UMIST Applied Mathematics, 1997.

\bibitem{kirby_hundley1999}
D.~Hundley and M.~Kirby,
\newblock ``Estimation of topological dimension,''
\newblock in {\em Proceedings of the Third SIAM International Conference on
  Data Mining}, San Fransico, 2001, pp. 194--202.

\bibitem{Hir94}
Morris~W. Hirsch,
\newblock {\em Differential topology}, vol.~33 of {\em Graduate Texts in
  Mathematics},
\newblock Springer-Verlag, New York, 1994,
\newblock Corrected reprint of the 1976 original.

\bibitem{AnHuKi02}
M.~Anderle, D.~Hundley, and M.~Kirby,
\newblock ``The bilipschitz criterion for mapping design in data analysis,''
\newblock {\em Intelligent Data Analysis}, vol. 6, no. 1, pp. 85--104, 2002.

\bibitem{kirby_wiley2}
M.~Kirby,
\newblock {\em Geometric Data Analysis: An Empirical Approach to Dimensionality
  Reduction and the Study of Patterns},
\newblock Wiley, 2001.

\bibitem{Fal03}
Kenneth Falconer,
\newblock {\em Fractal geometry},
\newblock John Wiley \& Sons, Inc., Hoboken, NJ, second edition, 2003,
\newblock Mathematical foundations and applications.

\bibitem{NBGS08}
John Nickolls, Ian Buck, Michael Garland, and Kevin Skadron,
\newblock ``Scalable parallel programming with {CUDA},''
\newblock {\em Queue}, vol. 6, no. 2, pp. 40--53, Mar. 2008.

\bibitem{cuSolver}
NVIDIA,
\newblock ``Cusolver library,''
\newblock 2018,
\newblock [Online; accessed 25-February-2018].

\bibitem{BK98}
Peter~N Belhumeur and David~J Kriegman,
\newblock ``What is the set of images of an object under all possible
  illumination conditions?,''
\newblock {\em International Journal of Computer Vision}, vol. 28, no. 3, pp.
  245--260, 1998.

\bibitem{GABK01}
Athinodoros~S. Georghiades, Peter~N. Belhumeur, and David~J. Kriegman,
\newblock ``From few to many: Illumination cone models for face recognition
  under variable lighting and pose,''
\newblock {\em IEEE transactions on pattern analysis and machine intelligence},
  vol. 23, no. 6, pp. 643--660, 2001.

\bibitem{CKKPDB07}
Jen-Mei Chang, Michael Kirby, Holger Kley, Chris Peterson, Bruce Draper, and
  J~Ross Beveridge,
\newblock ``Recognition of digital images of the human face at ultra low
  resolution via illumination spaces,''
\newblock in {\em Asian Conference on Computer Vision}. Springer, 2007, pp.
  733--743.

\bibitem{CBDKKP06}
Jen-Mei Chang, J~Ross Beveridge, Bruce~A Draper, Michael Kirby, Holger Kley,
  and Chris Peterson,
\newblock ``Illumination face spaces are idiosyncratic.,''
\newblock {\em IPCV}, vol. 2, pp. 390--396, 2006.

\bibitem{IP}
{Grupo de Inteligencia Computacional},
\newblock ``Hyperspectral remote sensing scences,'' 2014,
\newblock
  \url{http://www.ehu.eus/ccwintco/index.php/Hyperspectral_Remote_Sensing_Scenes},
  Last accessed on 2018-4-30.

\end{thebibliography}

\end{document}